\documentclass[runningheads]{llncs}
\usepackage[utf8]{inputenc}

\usepackage{hyperref}
\usepackage{xcolor}
\usepackage{lineno}
\usepackage{amsmath}
\usepackage{algpseudocode}
\usepackage{algorithm}
\usepackage{multirow}
\usepackage{graphicx}
\usepackage{todonotes}
\usepackage{booktabs}

\usepackage{float}
\newfloat{algorithm}{t}{lop}

\begin{document}

\title{On Monte Carlo Tree Search for Weighted Vertex Coloring}
\titlerunning{MCTS for weighted vertex coloring}

\author{Cyril Grelier \and Olivier Goudet \and Jin-Kao Hao\thanks{Corresponding author}}
\institute{LERIA, Université d’Angers, 2 Boulevard Lavoisier, 49045 Angers, France
\email{\{cyril.grelier,olivier.goudet,jin-kao.hao\}@univ-angers.fr}}
\authorrunning{Grelier, Goudet and Hao}

\maketitle

\begin{abstract}

This work presents the first study of using the popular Monte Carlo Tree Search (MCTS) method combined with dedicated heuristics for solving the Weighted Vertex Coloring Problem. Starting with the basic MCTS algorithm, we gradually introduce a number of algorithmic variants where MCTS is extended by various simulation strategies including greedy and local search heuristics. We conduct experiments on well-known benchmark instances to assess the value of each studied combination. We also provide empirical evidence to shed light on the advantages and limits of each strategy. 
   
\end{abstract}

\section{Introduction}
The well-known Graph Coloring Problem (GCP) is to color the vertices of a graph using as few colors as possible such that no adjacent vertices share the same color (\textit{legal} or \textit{feasible} solution). The GCP can also be considered as partitioning the vertex set of the graph into a minimum number of color groups such that no vertices in each color group are adjacent. The GCP has numerous practical applications in various domains \cite{Lewis16} and has been studied for a long time. 

A variant of the GCP called the Weighted Vertex Coloring Problem (WVCP) has recently attracted a lot of interest in the literature \cite{Goudet_Grelier_Hao_2021,Nogueira_Tavares_Maciel_2021,Sun_Hao_Lai_Wu_2018,Wang_Cai_Pan_Li_Yin_2020}. In this problem, each vertex of the graph has a weight and the objective is to find a \textit{legal} solution such that the sum of the weights of the heaviest vertex of each color group is minimized. Formally, given a weighted graph $G=(V,E)$ with vertex set $V$ and edge set $E$, and let $W$ be the set of weights $w(v)$ associated to each vertex $v$ in $V$, the WVCP consists in finding a partition of the vertices in $V$ into $k$ color groups $S=\{V_1,\dots,V_k\}$ ($1 \leq k \leq |V|)$ such that no adjacent vertices belong to the same color group and such that the score $\sum_{i=1}^{k}\max_{v\in V_i}{w(v)}$ is minimized. One can notice that when all the weights $w(v)$ ($v \in V$) are equal to one, finding an optimal solution of this problem with a minimum score corresponds to solving the GCP. Therefore, the WVCP can be seen as a more general problem than the GCP and is NP-hard. This extension of the GCP is useful for a number of applications in multiple fields such as matrix decomposition problem \cite{Prais_Ribeiro_2000}, buffer size management, scheduling of jobs in a multiprocessor environment \cite{Pemmaraju_Raman_2005} or partitioning a set of jobs into batches \cite{Kavitha_Mestre_2012}.

Different methods have been proposed in the literature to solve the WVCP. First, this problem has been tackled with exact methods: a branch-and-price algorithm \cite{furini2012exact} and two ILP models proposed in \cite{malaguti2009models} and \cite{Cornaz_Furini_Malaguti_2017}, which transforms the WVCP into a maximum weight independent set problem. These exact methods are able to prove the optimality on small instances, but fail on large instances. 

To handle large graphs, several heuristics have been introduced to solve the problem approximately \cite{Nogueira_Tavares_Maciel_2021,Prais_Ribeiro_2000,Sun_Hao_Lai_Wu_2018,Wang_Cai_Pan_Li_Yin_2020}. The first category of heuristics is local search algorithms, which iteratively make transitions from the current solution to a neighbor solution. Three different approaches have been considered to explore the search space: legal, partial legal, or penalty strategies. The legal strategy starts from a \textit{legal} solution and minimizes the score by performing only legal moves so that no color conflict is created in the new solution \cite{Prais_Ribeiro_2000}. The partial legal strategy allows only legal coloring and keeps a set of uncolored vertices to avoid conflicts \cite{Nogueira_Tavares_Maciel_2021}. The penalty strategy considers both legal and illegal solutions in the search space \cite{Sun_Hao_Lai_Wu_2018,Wang_Cai_Pan_Li_Yin_2020}, and uses a weighted evaluation function to minimize both the WVCP objective function and the number of conflicts in the illegal solutions. To escape local optima traps, these local search algorithms incorporate different mechanisms such as perturbation strategies \cite{Prais_Ribeiro_2000,Sun_Hao_Lai_Wu_2018}, tabu list \cite{Nogueira_Tavares_Maciel_2021,Sun_Hao_Lai_Wu_2018} and constraint reweighting schemes \cite{Wang_Cai_Pan_Li_Yin_2020}.
 
The second category of existing heuristics for the WVCP relies on the population-based memetic framework that combines local search with crossovers. A recent algorithm \cite{Goudet_Grelier_Hao_2021} of this category uses a deep neural network to learn an invariant by color permutation regression model, useful to select the most promising crossovers at each generation.
 
Research on combining such learning techniques and heuristics has received increasing attention in the past years \cite{zhou2018improving,zhou2020frequent}. In these new frameworks, useful information (e.g., relevant patterns) is learned from past search trajectories and used to guide a local search algorithm.

This study continues on this path and investigates the potential benefits of Monte Carlo Tree Search (MCTS) to improve the results of sequential coloring and local search algorithms. MCTS is a heuristic search algorithm that generated considerable interest due to its spectacular success for the game of Go \cite{gelly2006modification}, and in other domains (see a survey of \cite{browne2012survey} on this topic). It has been recently revisited in combination with modern deep learning techniques for difficult two-player games (cf. AlphaGo \cite{Silver_Huang_Maddison_2016}). MCTS has also been applied to combinatorial optimization problems seen as a one-player game such as the traveling salesman problem \cite{edelkamp2014solving} or the knapsack problem \cite{Jooken_Leyman_2020}. An algorithm based on MCTS has recently been implemented with some success for the GCP in \cite{Cazenave_Negrevergne_Sikora_2020}. In this work, we investigate for the first time the MCTS approach for solving the WVCP.

In MCTS, a tree is built incrementally and asymmetrically. For each iteration, a tree policy balancing exploration and exploitation is used to find the most critical node to expand. A simulation is then run from the expanded node and the search tree is updated with the result of this simulation. Its incremental and asymmetric properties make MCTS a promising candidate for the WVCP because in this problem only the heaviest vertex of each color group has an impact on the objective score. Therefore learning to color the heaviest vertices of the graph before coloring the rest of the graph seems particularly relevant for this problem. The contributions of this work are summarized as follows.

First, we present a new MCTS algorithm dedicated to the WVCP, which considers the problem from the perspective of a sequential coloring with a predefined vertex order. The exploration of the tree is accelerated with the use of specific pruning rules, which offer the possibility to explore the whole tree in a reasonable amount of time for small instances and to obtain optimality proofs. Secondly, for large instances, when obtaining an exact result is impossible in a reasonable time, we study how this MCTS algorithm can be tightly coupled with other heuristics. Specifically, we investigate the integration of different greedy coloring strategies and a local search procedure with the MCTS algorithm. 
 
The rest of the paper is organized as follows. Section \ref{sec:constructionTree} introduces the weighted vertex coloring problem and the constructive approach with a tree. Section \ref{sec:mcts_classic} describes the MCTS algorithm devised to tackle the problem. Section \ref{sec:mcts_improved} presents the coupling of  MCTS with a local search algorithm. Section \ref{sec:experimentation} reports results of the different versions of MCTS. Section \ref{sec:conclusion} discusses the contributions and presents research perspectives.
 
\section{Constructive approach with a tree for the weighted graph coloring problem \label{sec:constructionTree}}

This section presents a tree-based approach for the WVCP, which aims to explore the partial and legal search space of this problem. 

\subsection{Partial and legal search space \label{sec:search_space}}

The search space $\Omega$ studied in our algorithm concerns legal, but potentially partial, $k$-colorings. A partial legal $k$-coloring $S$ is a partition of the set of vertices $V$ into $k$ disjoint independent sets $V_i$ ($1 \leq i \leq k$), and a set of uncolored vertices $U = V \backslash \bigcup_{i=1}^{k}V_i$. A independent set $V_i$ is a set of mutually non adjacent vertices of the graph: $\forall u,v \in V_i, (u,v) \notin E$. For the WVCP, the number of colors $k$ that can be used is not known in advance. Nevertheless, it is not lower than the chromatic number of the graph $\chi(G)$ and not greater than the number of vertices $|V|$ of the graph. A solution of the WVCP is denoted as partial if $U \neq \emptyset$ and complete otherwise. The objective of the WVCP is to find a complete solution $S$ with a minimum score $f(S)$ given by: $f(S) = \sum_{i=1}^{k}\max_{v\in V_i}{w(v)}.$

\subsection{Tree search for weighted vertex coloring} \label{sec:treeSearch}

Backtracking based tree search is a popular approach for the graph coloring problem \cite{brelaz1979new,KubaleJ84,Lewis16}. In our case, a tree search algorithm can be used to explore the partial and legal search space of the WVCP previously defined. 

Starting from a solution where no vertex is colored (i.e., $U=V$) and that corresponds to the root node $R$ of the tree, child nodes $C$ are successively selected in the tree, consisting of coloring one new vertex at a time. This process is repeated until a terminal node $T$ is reached (all the vertices are colored). A complete solution (i.e., a legal coloring) corresponds thus to a branch from the root node to a terminal node. 

The selection of each child node corresponds to applying a move to the current partial solution being constructed. A move consists of assigning a particular color $i$ to an uncolored vertex $u \in U$, denoted as $<u,U,V_i>$. Applying a move to the current partial solution $S$, results in a new solution $S\ \oplus <u,U,V_i>$. This tree search algorithm only considers legal moves to stay in the partial legal space. For a partial solution $S=\{V_1,...,V_k,U\}$, a move $<u,U,V_i>$ is said legal if no vertex of $V_i$ is adjacent to the vertex $u$. At each level of the tree, there is at least one possible legal move that applies to a vertex a new color that has never been used before (or putting this vertex in a new empty set $V_{i}$, with $k+1 \leq i \leq |V|$). 

Applying a succession of $|V|$ legal moves from the initial solution results in a legal coloring of the WVCP and reaches a terminal node of the tree. During this process, at the level $t$ of the tree ($0 \leq t < |V|)$, the current legal and partial solution $S=\{V_1,...,V_k,U\}$ has already used $k$ colors and $t$ vertices have already received a color. Therefore $|U| = |V| - t$.

At this level, a first naive approach could be to consider all the possible legal moves, corresponding to choosing a vertex in the set $U$ and assigning to the vertex a color $i$, with $1 \leq i \leq |V|$. This kind of choice can work with small graphs but with large graphs, the number of possible legal moves becomes huge. Indeed, at each level $t$, the number of possible legal moves can go up to $(|V| - t) \times |V|$. To reduce the set of move possibilities, we consider the vertices of the graph in a predefined order $(v_1,\dots,v_{|V|})$. Moreover, to choose a color for the incoming vertex, we consider only the colors already used in the partial solution plus one color (creation of a new independent set). Thus for a current legal and partial solution $S=\{V_1,...,V_k,U\}$, at most $k+1$ moves are considered. The set of legal moves is:
\begin{equation}
 \mathcal{L}(S) = \{ <u,U,V_i>, 1 \leq i \leq k, \forall v \in V_i, (u,v) \notin E \}\ \cup <u,U,V_{k+1}> \label{eq:set_legal_moves}
\end{equation}
This decision cuts the symmetries in the tree while reducing the number of branching factors at each level of the tree.

\subsection{Predefined vertex order}\label{sec:order}

We propose to consider a predefined ordering of the vertices, sorted by weight then by degree. Vertices with higher weights are placed first. If two vertices have the same weight, then the vertex with the higher degree is placed first. This order is intuitively relevant for the WVCP, because it is more important to place first the vertices with heavy weights which have the most impact on the score as well as the vertices with the highest degree because they are the most constrained decision variables. Such ordering has already been shown to be effective with greedy constructive approaches for the GCP \cite{brelaz1979new} and the WVCP \cite{Nogueira_Tavares_Maciel_2021}.
 
Moreover, this vertex ordering allows a simple score calculation while building the tree. Indeed, as the vertices are sorted by descending order of their weights, and the score of the WVCP only counts the maximum weight of each color group, with this vertex ordering, the score only increases by the value $w(v)$ when a new color group is created for the vertex $v$.

\section{Monte Carlo Tree Search for weighted vertex coloring} \label{sec:mcts_classic}

The search tree presented in the last subsection can be huge, in particular for large instances. Therefore, in practice, it is often impossible to perform an exhaustive search of this tree, due to expensive computing time and memory requirements. We turn now to an adaptation of the MCTS algorithm for the WVCP to explore this search tree. MCTS keeps in memory a tree (hereinafter referred to as the MCTS tree) which only corresponds to the already explored nodes of the search tree presented in the last subsection. In the MCTS tree, a leaf is a node whose children have not yet all been explored while a terminal node corresponds to a complete solution.  MCTS can guide the search toward the most promising branches of the tree, by balancing exploitation and exploration and continuously learning at each iteration. 

\subsection{General framework}

The MCTS algorithm for the WVCP is shown in Algorithm \ref{alg:MCTS}. The algorithm takes a weighted graph as input and tries to find a legal coloring $S$ with the minimum score $f(S)$. The algorithm starts with an initial solution where the first vertex is placed in the first color group. This is the root node of the MCTS tree. Then, the algorithm repeats a number of iterations until a stopping criterion is met. At every iteration, one legal solution is completely built, which corresponds to walking along a path from the root node to a leaf node of the MCTS tree and performing a simulation (or playout/rollout) until a terminal node of the search tree is reached (when all vertices are colored).

Each iteration of the MCTS algorithm involves the execution of 5 steps to explore the search tree with legal moves (cf. Section \ref{sec:constructionTree}):

\begin{enumerate}
    \item \textbf{Selection} From the root node of the MCTS tree, successive child nodes are selected until a leaf node is reached. The selection process balances the exploration-exploitation trade-off. The exploitation score is linked to the average score obtained after having selected this child node and is used to guide the algorithm to a part of the tree where the scores are the lowest (the WVCP is a minimization problem). The exploration score is linked to the number of visits to the child node and will incite the algorithm to explore new parts of the tree, which have not yet been explored.
    \item \textbf{Expansion} The MCTS tree grows by adding a new child node to the leaf node reached during the selection phase.
    \item \textbf{Simulation} From the newly added node, the current partial solution is completed with legal moves, randomly or by using heuristics.
    \item \textbf{Update} After the simulation, the average score and the number of visits of each node on the explored branch are updated.
    \item \textbf{Pruning} If a new best score is found, some branches of the MCTS tree may be pruned if it is not possible to improve the best current score with it.
\end{enumerate}

\noindent The algorithm continues until one of the following conditions is reached: 
\begin{itemize}
 \item there are no more child nodes to expand, meaning the search tree has been fully explored. In this case, the best score found is proven to be optimal.
 \item a cutoff time is attained. The minimum score found so far is returned. It corresponds to an upper bound of the score for the given instance.
\end{itemize}

\begin{algorithm} \caption{MCTS algorithm for the WVCP\label{alg:MCTS}}
    \begin{algorithmic}[1]
        \State \bf{Input}: \normalfont{Weighted graph $G = (V, W, E)$}
        \State \bf{Output}: \normalfont{The best legal coloring $S^*$ found}
        \State $S^*=\emptyset$ and $f(S^*) = MaxInt$
        \While{stop condition is not met}
            \State $C \leftarrow R$ \Comment{Current node corresponding to the root node of the tree}
            \State $S$ $\leftarrow$ $\{V_1,U\}$ with $V_1 = \{v_1\}$ and $U = V \backslash V_1$ \Comment{Current solution initialized with the first vertex in the first color group}
            \State 
            /* Selection */ \Comment{Section \ref{mcts:selection}}
            \While{$C$ is not a leaf}
                \State $C$ $\leftarrow$ select\_best\_child(C)  with legal move $<u,U,V_i>$
                \State $S \leftarrow S\ \oplus <u,U,V_i>$ 
            \EndWhile
            \State 
            /* Expansion */ \Comment{Section \ref{mcts:expansion}}
            \If{C has a potential child, not yet open} 
                \State $C$ $\leftarrow$ open\_first\_child\_not\_open(C) with legal move $<u,U,V_i>$
                \State $S \leftarrow S\ \oplus <u,U,V_i>$ 
            \EndIf
            \State 
            /* Simulation */ \Comment{Section \ref{mcts:simulation}}
            \State complete\_partial\_solution($S$)
            \State 
            /* Update */ \Comment{Section \ref{mcts:update}}
            \While{$C \neq R$}
                \State update(C,f(S))
                \State $C \leftarrow$ parent(C)
            \EndWhile
            \If{$f(S) < f(S^*)$}
                \State $S^* \leftarrow S$
                \State /* Pruning */ \Comment{Section \ref{mcts:pruning}}
                \State apply pruning rules
            \EndIf
        \EndWhile
        \State return $S^*$
    \end{algorithmic}
\end{algorithm}

\subsection{Selection} \label{mcts:selection}

The selection starts from the root node of the MCTS tree and selects children nodes until a leaf node is reached. At every level $t$ of the MCTS tree, if the current node $C_t$ is corresponds to a partial solution $S=\{V_1,...,V_k,U\}$ with $t$ vertices already colored and $k$ colors used, there are $l$ possible legal moves, with $1 \leq l \leq k+1$. Therefore, from the node $C_t$, $l$ potential children $C^1_{t+1}, \dots, C^l_{t+1}$ can be selected. 

If $l > 1$, the selection of the most promising child node can be seen as a multi-armed bandit problem \cite{Lai_Robbins_1985} with $l$ levers. This problem of choosing the next node can be solved with the UCT algorithm for Monte Carlo tree search by selecting the child with the maximum value of the following expression \cite{Jooken_Leyman_2020}: 
\begin{equation}
 normalized\_score(C^i_{t+1}) + c \times \sqrt{\frac{2* ln(nb\_visits(C_t))}{nb\_visits(C^i_{t+1})}},\ \text{for}\ 1 \leq i \leq l.
 \label{eq:score_UCT}
\end{equation}
\noindent Here, $nb\_visits(C)$ corresponds to the number of times the node $C$ has been chosen to build a solution. $c$ is a real positive coefficient allowing to balance the compromise between exploitation and exploration. It is set by default to the value of one. $normalized\_score(C^i_{t+1})$ corresponds to a normalized score of the child node $C^i_{t+1}$ ($1 \leq i \leq l$) given by:
$$normalized\_score(C^i_{t+1})= \frac{rank(C^i_{t+1})}{\sum_{i=1}^l rank(C^i_{t+1})},$$

\noindent where $rank(C^i_{t+1})$ is defined as the rank between 1 and $l$ of the nodes $C^i_{t+1}$ obtained by sorting from bad to good according their average values $avg\_score(C^i_{t+1})$ (nodes that seem more promising get a higher score). $avg\_score(C^i_{t+1})$ is the mean score on the sub-branch with the node $C^i_{t+1}$ selected obtained after all previous simulations.

\subsection{Expansion} \label{mcts:expansion}

From the node $C$ of the MCTS tree reached during the selection procedure, one new child of $C$ is open and its corresponding legal move is applied to the current solution.  Among the unopened children, we select the node associated with the lowest color number $i$. Therefore the child node needing the creation of a new color (and increasing the score) will be selected last.

\subsection{Simulation} \label{mcts:simulation}

The simulation takes the current partial and legal solution found after the expansion phase and colors the remaining vertices. In the original MCTS algorithm, the simulation consists in choosing random moves in the set of all legal moves $\mathcal{L}(S)$ defined by equation (\ref{eq:set_legal_moves}) until the solution is completed. We call this first version \textit{random-MCTS}. As shown in the experimental section, this version is not very efficient as the number of colors grows rapidly. Therefore, we propose two other simulation procedures: 
\begin{itemize}
    \item a constrained greedy algorithm that chooses a legal move prioritizing the moves which do not locally increase the score of the current partial solution $S=\{V_1,...,V_k,U\}$: 
    \begin{equation}
        \mathcal{L}^g(S) = \{ <u,U,V_i>, 1 \leq i \leq k, \forall v \in V_i, (u,v) \notin E \}
        \label{eq:set_legal_moves_no_new_color}
    \end{equation}
    It only chooses the move $<u,U,V_{k+1}>$, consisting in opening a new color group and increasing the current score by $w(u)$, only if $\mathcal{L}^g(S) = \emptyset$. We call this version \textit{greedy-random-MCTS}.
    \item a greedy deterministic procedure which always chooses a legal move in $\mathcal{L}(S)$ with the first available color $i$. We call this version \textit{greedy-MCTS}.
        
\end{itemize}

\subsection{Update} \label{mcts:update}

Once the simulation is over, a complete solution of the WVCP is obtained. If this solution is better than the best recorded solution found so far $S^*$ (i.e., $f(S) < f(S^*)$), $S$ becomes the new global best solution $S^*$.

Then, a backpropagation procedure updates each node $C$ of the whole branch of the MCTS tree which has led to this solution:
\begin{itemize}
    \item the running average score of each node $C$ of the branch is updated with the score $f(S)$:
    \begin{equation}
        avg\_score(C) \leftarrow \frac{avg\_score(C) \times nb\_visits(C) + f(S)}{nb\_visits(C) + 1} 
    \end{equation}
    \item the counter of visits $nb\_visits(C)$ of each node of the branch is increased by one.
\end{itemize}

\subsection{Pruning} \label{mcts:pruning}

During an iteration of MCTS, three pruning rules are applied:

\begin{enumerate}
    \item during expansion: if the score $f(S)$ of the partial solution associated with a node visited during this iteration of MCTS is equal or higher to the current best-found score $f(S^*)$, then the node is deleted as the score of a partial solution cannot decrease when more vertices are colored.
    \item when the best score $f(S^*)$ is found, the tree is \textit{cleaned}. A heuristic goes through the whole tree and deletes children and possible children associated with a partial score $f(S)$ equal or superior to the best score $f(S^*)$. 
    \item if a node is \textit{completely explored}, it is deleted and will not be explored in the MCTS tree anymore. A node is said \textit{completely explored} if it is a leaf node without children, or if all of its children have already been opened once and have all been deleted. Note that this third pruning step is recursive as a node deletion can result in the deletion of its parent if it has no more children, and so on.
\end{enumerate}

These three pruning rules and the fact that the symmetries are cut in the tree by restricting the set of legal moves considered at each step (see Section \ref{sec:treeSearch}) offer the possibility to explore the whole tree in a reasonable amount of time for small instances. This peculiarity of the algorithm makes it possible to obtain an optimality proof for such instances.

\section{Improving MCTS with Local Search} \label{sec:mcts_improved}

We now explore the possibility of improving the MCTS algorithm with local search. Coupling MCTS with a local search algorithm is motivated by the fact that after the simulation phase, the complete solution obtained can be close to a still better solution in the search space that could be discovered by local search. 

For this purpose, we propose a strong integration between the MCTS framework and a local search procedure regarding two aspects:
\begin{itemize}
    \item first, to stay consistent with the search tree learned by MCTS, we allow the local search procedure to only move the vertices of the complete solution $S$ which are still uncolored after the selection and expansion phases. This set of free vertices is called $F_V$. 
    \item secondly, the best score obtained by the local search procedure is used to update all the nodes of the branch which has led to the simulation initiation.
\end{itemize}

When coupling MCTS with a local search algorithm, the resulting heuristic can be seen as an algorithm that attempts to learn a good starting point for the local search procedure, by selecting a different path in every iteration during the selection phase.

In this work, we present the coupling of MCTS with an iterated tabu search (ITS) that searches in the same legal and partial search space presented in Section \ref{sec:constructionTree}. We call this variant \textit{MCTS+ITS}. The ITS algorithm is an adaptation of the ILS-TS algorithm \cite{Nogueira_Tavares_Maciel_2021}, which is one of the best performing local search algorithms for the WVCP. 
 
 From a complete solution, the ITS algorithm iteratively performs 2 steps:
\begin{itemize}
    \item first it deletes the heaviest vertices in $F_V$ from 1 to 3 color groups $V_i$ and places them in the set of uncolored vertices $U$.
    \item then, it improves the solution (i.e., minimizes the score $f(S)$) by applying the \textit{one move} operator (consisting in moving a vertex from one color group to another) and the so-called \textit{grenade} operator until the set of uncolored vertices $U$ becomes empty.
\end{itemize}

The grenade operator $grenade(u,V_i)$ consists in moving a vertex $u$ to $V_i$, but first, each adjacent vertice of $u$ in $V_i$ is relocated to other color groups or in $U$ to keep a legal solution. In our algorithm, as we restrict the vertices allowed to move to the set of free vertices $F_V$, the neighborhood of a solution $S$ associated with this grenade operator is given by: $N(S) = \{S\ \oplus\ grenade(u,V_i): u \in F_V, u \notin V_i, 1 \leq i \leq k, \forall v \in V_i, \left((u,v) \notin E\ or\ v \in F_V \right)\}$.

\section{Experimentation} \label{sec:experimentation}
		
This section aims to experimentally verify the impacts of the different greedy coloring strategies used during the MCTS simulation phase, and the interest of coupling MCTS with local search. 

\subsection{Experimental settings and benchmark instances}

A total of 161 instances in the literature were used for the experimental studies: 30 rxx graphs and 35 pxx graphs from matrix decomposition \cite{Prais_Ribeiro_2000} and 46 small and 50 large graphs from the DIMACS and COLOR competitions.

Two preprocessing procedures were applied to reduce the graphs of the different instances. The first one comes from \cite{Wang_Cai_Pan_Li_Yin_2020}: if the weight of a vertex of degree $d$ is lower than the weight of the $d+1$th heaviest vertex from any clique of the graph, then the vertex can be deleted without changing the optimal WVCP score of this instance. The second one comes from \cite{Cheeseman_Kanefsky_Taylor_1991} and is adapted to our problem: if all neighbors of a vertex $v1$ are all neighbors with a vertex $v2$ and the weight of $v2$ is greater or equal to the weight of $v1$, then $v1$ can be deleted as it can take the color of $v2$ without impacting the score.

The MCTS algorithms of Section \ref{sec:mcts_classic} were coded in C++ and compiled with the g++ 8.3 compiler and the -O3 optimization flag using GNU Parallel \cite{Tange2011a}. To solve each instance, 20 independent runs were performed on a computer equipped with an Intel Xeon E7-4850 v4 CPU with a time limit of one hour. Running the DIMACS Machine Benchmark procedure dfmax\footnote{\url{http://archive.dimacs.rutgers.edu/pub/dsj/clique/}} on our computer took 9 seconds to solve the instance r500.5 using gcc 8.3 without optimization flag. The source code of our algorithm is available at \url{https://github.com/Cyril-Grelier/gc_wvcp_mcts}. 

For our experimental studies, we report the following statistics in the tables of results. Column BKS (Best-Known Score) reports the best scores from all methods reported in the literature which have been obtained using different computing tools, CPU or GPU (Graphics Processing Units), sometimes under relaxed conditions (specific fine-tuning of the hyperparameters, large run time up to several days). The scores of the different methods equal to this best-known score are marked in bold in the tables. The optimal results obtained with the exact algorithm MWSS \cite{Cornaz_Furini_Malaguti_2017} and reported in \cite{Nogueira_Tavares_Maciel_2021} (launched with a time limit of ten hours for each instance) are indicated with a star. Column "best" shows the best results for a method, "Avg" the average score on the 20 runs, and "t(s)" the average time in seconds to reach the best scores. 

Due to space limitations, we only present the results for 25 out of the 161 instances tested. These 25 instances can be considered to be among the most difficult ones.

\subsection{Monte Carlo Tree Search with greedy strategies}

Table \ref{tab:mcts_classic} shows the results of MCTS with the greedy heuristics for its simulation (cf. Section \ref{mcts:simulation}). Column 2 gives the number of vertices of the graph. Column 3 shows the best-known score ever reported in the literature for each instance. Column 4 corresponds to the baseline greedy algorithm: using the predefined vertex ordering presented in Section \ref{sec:order}, each vertex is placed in the first color group available without creating conflicts (if no color is available, a new color group is created). Columns 5-13 show the results of the three greedy strategies used in the simulation phase of MCTS (cf. Section \ref{mcts:simulation}): random-MCTS, greedy-random-MCTS, and greedy-MCTS. Column 14 presents the gap between the mean score of the greedy-random-MCTS and the greedy-MCTS variants. Significant gaps (t-test with a p-value below 0.001) are underlined.

\begin{table}[h]
\centering
\caption{Comparison between the random-MCTS, greedy-random-MCTS and greedy-MCTS variants (\textbf{best known}, proven optimum*, \underline{significant gap})}
\label{tab:mcts_classic}
\resizebox{\textwidth}{!}{
\begin{tabular}{c|c|c|c|ccc|ccc|ccc|c}\hline
                &  & BKS   & greedy & \multicolumn{3}{c|}{random-MCTS (1)}       & \multicolumn{3}{c|}{greedy-random-MCTS (2)}   & \multicolumn{3}{c|}{greedy-MCTS (3)} & gap     \\
instance & $|V|$  & best  & best         & best         & avg     & t(s)  & best          & avg     & t(s)    & best          & avg    & t(s)    & (3) - (2) \\  \hline
C2000.5 & 2000    & 2151  & 2470         & 3262         & 3281.5  & 3150  & 2584          & 2603.6  & 2524    & 2386 & 2387.7 & 3462    & \underline{-215.9}    \\
C2000.9 & 2000    & 5486  & 6449         & 7300         & 7334.4  & 1128  & 6420          & 6452.3  & 3403    & 6249 & 6249   & 382.6   & \underline{-203.3}    \\
DSJC500.1 & 500  & 184   & 224          & 240          & 247.6   & 1243  & 205  & 208.15  & 358     & 209           & 209    & 480.6   & 0.85       \\
DSJC500.5 & 500  & 686   & 840          & 815          & 827.7   & 1336  & 760           & 766.5   & 606     & 757  & 757    & 1037.1  & \underline{-9.5}      \\
DSJC500.9 & 500   & 1662  & 1916         & 1830         & 1857.5  & 680   & 1766 & 1788.3  & 496     & 1788          & 1788   & 1354.65 & -0.3         \\ \hline
DSJR500.1 & 500   & 169   & 187          & 179          & 184.9   & 656   & \textbf{169}  & 169.05  & 496.05  & 177           & 177    & 4.6     & \underline{7.95}    \\
flat1000\_50\_0 & 1000  & 924   & 1350   & 1572          & 1639.7  & 3211   & 1284          & 1304.9  & 3566    & 1273 & 1273.6  & 3516    & \underline{-31.3} \\
flat1000\_60\_0 & 1000  & 1162  & 1388   & 1666          & 1700.9  & 3571   & 1330          & 1346.75 & 3416    & 1303 & 1303.15 & 3265.26 & \underline{-43.6} \\
GEOM120b & 120    & 35*   & 40           & 38           & 38.55   & 42.69 & 36   & 36.95   & 4       & 38            & 38     & 0       & \underline{1.05}    \\
inithx.i.1 & 864  & 569   & \textbf{569} & 571          & 574.35  & 3584  & \textbf{569}  & 569     & 0       & \textbf{569}  & 569    & 0       & 0       \\ \hline
latin\_square\_10 & 900 & 1483  & 1876   & 2178          & 2288.15 & 3575   & 1729 & 1766.25 & 3444    & 1766          & 1766    & 2746.85 & -0.25    \\
le450\_25a & 450  & 306   & 321          & 313          & 318.35  & 2311  & 308  & 313.45  & 3380    & 312           & 312    & 18.7    & -1.45       \\
le450\_25b & 450       & 307   & 312    & 310           & 312.7   & 2217.5 & 309  & 309.4   & 1797.85 & 309  & 309     & 0       & -0.4  \\
miles1500 & 128   & 797*  & 799          & \textbf{797} & 797     & 1.3   & \textbf{797}  & 797     & 0       & \textbf{797}  & 797    & 0       & 0       \\
p41     &   116       & 2688* & 2724   & \textbf{2688} & 2709.55 & 446.17 & \textbf{2688} & 2688    & 0.15    & \textbf{2688} & 2688    & 0       & 0    \\ \hline
p42   &  138     & 2466* & 2517         & 2489         & 2514.05 & 35    & \textbf{2466} & 2466.45 & 10      & \textbf{2466} & 2466   & 0.95    & -0.45        \\
queen10\_10 & 100 & 162   & 190          & 173          & 178.2   & 691.5 & 164  & 168.55  & 1       & 169           & 169    & 7.35    & 0.45       \\
queen16\_16 & 256 & 234   & 264          & 269          & 274.6   & 854   & 248           & 252.65  & 116     & 244  & 244    & 70.25   & \underline{-8.65}     \\
r28   &  288     & 9407* & 9409         & 9439         & 9560.8  & 543   & \textbf{9407} & 9411.9  & 2413.8  & \textbf{9407} & 9407   & 0       & \underline{-4.9}      \\
r29   &  281     & 8693* & 8973         & 8875         & 9054    & 1519  & \textbf{8693} & 8698.05 & 1723.5  & 8694          & 8694   & 0.2     & \underline{-4.05}     \\ \hline
r30  & 301       & 9816* & 9831         & 9826         & 9881.4  & 770   & \textbf{9816} & 9819.6  & 27.75   & 9818          & 9818   & 0       & -1.6         \\
R50\_5gb &  50        & 135*  & 158    & \textbf{135*}  & 135     & 727    & \textbf{135*}  & 135     & 652.45  & \textbf{135*}  & 135     & 384.4   & 0    \\
R75\_1gb & 70   & 70*   & 91           & 74           & 78.4    & 170   & \textbf{70*}   & 74.05   & 1150    & 72            & 72     & 1291.1  & \underline{-2.05}     \\
wap01a  & 2368    & 545   & 628          & 1050         & 1071.95 & 3530  & 659           & 663.05  & 2278    & 603  & 603    & 1385.9  & \underline{-60.05}    \\
wap02a &  2464    & 538   & 619          & 1034         & 1051.2  & 3505  & 651           & 653.6   & 1268.67 & 591  & 591    & 1586.05 & \underline{-62.6}     \\ \hline
     &       &       & 1/25         & \multicolumn{3}{c|}{3/25}       & \multicolumn{3}{c|}{10/25}         & \multicolumn{3}{c|}{6/25}        &  \\
      \hline
\end{tabular}
}
\end{table}

First, we observe that the version random-MCTS (Columns 5-7) performs badly even in comparison with the baseline greedy algorithm (Column 4). This is particularly true for the largest instances such as C2000.5 and C2000.9. Indeed, for large instances, choosing random moves in the set of all legal moves is not very efficient as the number of color groups grows rapidly. 

The greedy-random-MCTS variant (Columns 8-10) is equipped with a simulation procedure that chooses a legal move in priority in the set of all possible legal moves which does not locally increase the WVCP score. We observe that this simulation strategy really improves the scores of MCTS in comparison with the random-MCTS version (Columns 5-7) and can reach 10 out of the 25 best-known results. 
This better heuristic used during the simulation phase helped also to prove the optimality for the instance R75\_1gb (indicated with a star) because the best score of 70 was found during the search which led to a pruning of the tree that allowed the tree to be fully explored earlier.

The greedy-MCTS variant (Columns 11-13) always dominates the baseline greedy algorithm (Column 3) for all the instances as it always starts with a simulation using the greedy algorithm at the root node of the tree (first simulation). It only finds 6 over 25 best-known results but is better for the largest instance when we compare the average results with those obtained by the greedy-random-MCTS variant ((Column 14). This is probably due to the fact that forcing the heaviest vertices to be grouped together in the first colors enables a better organization of the color groups. This greedy variant used during the simulation leads to a better intensification which is beneficial for large instances, for which the search tree cannot be sufficiently explored due to the time limit of one hour. 

However, with the deterministic simulation of this greedy-MCTS variant, we lose the sampling part with the random legal moves used in the greedy-random-MCTS allowing greater exploration of the search space and a better estimation of the most promising branches of the search tree.
This particularity of the greedy-random-MCTS allows finding the best-known score for DSJR500.1, r29, or R75\_1gb for example. 

\begin{figure}[!h]
\centering
    \includegraphics[scale=0.35]{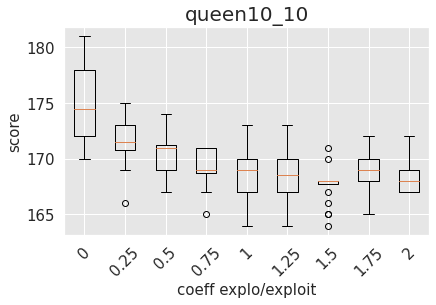}
    \includegraphics[scale=0.35]{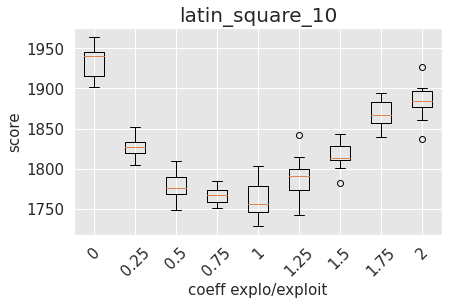}
    \includegraphics[scale=0.35]{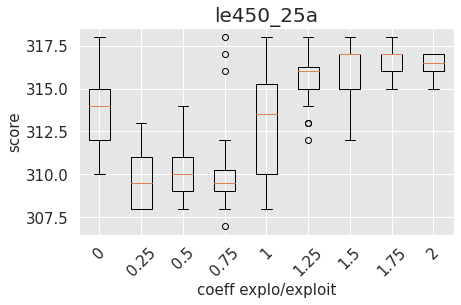}
    \includegraphics[scale=0.35]{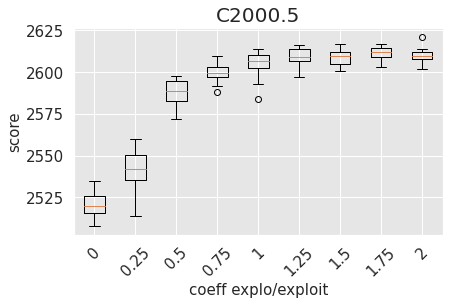}
    \caption{Boxplots of the distribution of the scores for different values of the coefficient $c$ between 0 and 2, for the instances queen10\_10, le450\_25c, latin\_square\_10 and C2000.5. For each configuration, 20 runs are launched with the greedy-random-MCTS variant.}
    \label{fig:coeff}
\end{figure}

To confirm this intuition, we varied the coefficient $c$ balancing the compromise between exploration and exploitation in equation \ref{eq:score_UCT} from 0 (no exploration) to 2 (encourage exploration) for different instances. For each value of the coefficient, we performed 20 runs of greedy-random-MCTS per instance during 1h per run. Figure \ref{fig:coeff} displays 4 box plots showing the distribution of the best scores for the instances queen10\_10, le450\_25c, latin\_square\_10 and C2000.5 for the different values of $c$. We observe 3 typical patterns also seen for the other instances. 

The pattern observed for the instance queen10\_10 is typical for easy instances or medium instances. For these instances, having no exploration performs badly, then the results get better when the coefficient rises to approximately 1 where it starts to stagnate. For these instances, having no exploration rapidly leads to a local minimum trap and it seems better to secure a minimum of diversity to reach a better score.

Conversely, for very large instances such as C2000.5, we observe a pattern completely different with a best solution when $c=0$. Indeed, for very large instances, as the search tree is huge and cannot be sufficiently explored due to the time limit, it seems more beneficial for the algorithm to favor more intensification to better search for a good solution in a small part of the tree.

An intermediate pattern appears for medium/hard instances such as le450\_25c or latin\_square\_10 with a U-shaped curve. For these instances, an optimal value of the coefficient $c$ is around the value of 1, with a good balance between exploitation and exploration. 

\subsection{Monte Carlo Tree Search with local search}

Table \ref{tab:mcts_improved_ilsts} reports the results of the greedy-MCTS variant (Columns 4-6), the state-of-the art local search ILS-TS algorithm \cite{Nogueira_Tavares_Maciel_2021} (Columns 7-9) and the MCTS algorithm combined with ITS (MCTS+ITS) with 500 iterations per simulation (Columns 10-12). Column 13 shows the gap between the mean scores of greedy-MCTS and MCTS+ITS. Column 14 shows the gap between the mean scores of ILS-TS and MCTS+ITS. Statistically significant gaps (t-test with a p-value below 0.001) are underlined.

As shown in Table \ref{tab:mcts_improved_ilsts}, the combination MCTS+ITS significantly improves the results of the greedy-MCTS version (without local search). However, the advantage of this coupling is less convincing compared with the results of the ILS-TS algorithm alone. In fact it degrades the results significantly for 2 instances DSJC500.9 and C2000.9 on average. Still, it greatly improves the average results for the flat1000\_60\_0 instance.

This may be due to the size of the tree that grows too fast for large instances with a high edge density such as DSJC500.9 and C2000.9. The algorithm does not allow to intensify the search enough with a timeout of 1h. This method would probably require a much longer time to learn more information on the higher part of the tree to reach better results, in particular for large instances.

We also studied combinations of MCTS with other local searches recently proposed in the literature for this problem such as AFISA \cite{Sun_Hao_Lai_Wu_2018} or RedLS \cite{Wang_Cai_Pan_Li_Yin_2020} (these results are available at \url{https://github.com/Cyril-Grelier/gc_wvcp_mcts}). The combination of MCTS with these methods significantly improves the results compared to the local search alone. In general, we observe that the weaker the local search, the greater the interest of combining it with MCTS. ILS-TS \cite{Nogueira_Tavares_Maciel_2021} being a powerful local search algorithm, the utility of combining it with the MCTS is actually less obvious.

\begin{table}[h!]
\centering
\caption{Comparison between the greedy-MCTS variant, ILS-TS \cite{Nogueira_Tavares_Maciel_2021}, and MCTS + ITS. (\textbf{best known}, proven optimal*, \underline{significant gap})}
\label{tab:mcts_improved_ilsts}
\resizebox{\textwidth}{!}{
\begin{tabular}{c|c|c|ccc|ccc|ccc|c|c}\hline
&
 &
  BKS &
  \multicolumn{3}{c|}{greedy-MCTS (1)} &
  \multicolumn{3}{c|}{ILS-TS (2)} &
  \multicolumn{3}{c|}{MCTS+ITS (3)} &
  gap &
  gap \\
instance &
$|V|$ &
  best &
  best &
  avg &
  t(s) &
  best &
  avg &
  t(s) &
  best &
  avg &
  t(s) &
  (3) - (1) &
  (3) – (2) \\ \hline
C2000.5 &
2000 &
  2151 &
  2386 &
  2387.7 &
  3462 &
  2256 &
  2278.52 &
  3438.27 &
  2251 &
  2273.45 &
  3658 &
  \underline{-114.25} &
  -5.07 \\
C2000.9 &
2000 &
  5486 &
  6249 &
  6249 &
  382.6 &
  5944 &
  5996.76 &
  3579.44 &
  6056 &
  6105.4 &
  3611 &
  \underline{-143.6} &
  \underline{108.64} \\
DSJC500.1 &
500 &
  184 &
  209 &
  209 &
  480.6 &
  185 &
  188.52 &
  3226.7 &
  189 &
  189.35 &
  1485.08 &
  \underline{-19.65} &
  0.83 \\
DSJC500.5 &
500 &
  686 &
  757 &
  757 &
  1037.1 &
  719 &
  735.76 &
  320.75 &
  721 &
  737.9 &
  2124 &
  \underline{-19.1} &
  2.14 \\
DSJC500.9 &
500 &
  1662 &
  1788 &
  1788 &
  1354.65 &
  1711 &
  1726.1 &
  2178.08 &
  1728 &
  1755.45 &
  2900 &
  \underline{-32.55} &
  \underline{29.35} \\ \hline
DSJR500.1 &
500 &
  169 &
  177 &
  177 &
  4.6 &
  \textbf{169} &
  169 &
  0.34 &
  \textbf{169} &
  169 &
  31.75 &
  \underline{-8} &
  0 \\
flat1000\_50\_0 &
1000 &
  924 &
  1273 &
  1273.6 &
  3516 &
  1220 &
  1235 &
  192.4 &
  1211 &
  1228.7 &
  2555 &
  \underline{-44.9} &
  -6.3 \\
flat1000\_60\_0 &
1000 &
  1162 &
  1303 &
  1303.15 &
  3265.26 &
  1252 &
  1271.86 &
  283.38 &
  1248 &
  1261.05 &
  2787 &
  \underline{-42.1} &
  \underline{-10.81} \\
GEOM120b &
120 &
  35* &
  38 &
  38 &
  0 &
  \textbf{35} &
  35 &
  1.06 &
  \textbf{35} &
  35 &
  7.05 &
  \underline{-3} &
  0 \\
inithx.i.1 &
864 &
  569 &
  \textbf{569} &
  569 &
  0 &
  \textbf{569} &
  569 &
  0 &
  \textbf{569} &
  569 &
  59.45 &
  0 &
  0 \\ \hline
latin\_square\_10 &
900 &
  1483 &
  1766 &
  1766 &
  2746.85 &
  1547 &
  1577.14 &
  2508.73 &
  1566 &
  1591.1 &
  3613 &
  \underline{-174.9} &
  13.96 \\
le450\_25a &
450 &
  306 &
  312 &
  312 &
  18.7 &
  \textbf{306} &
  306.29 &
  679.64 &
  \textbf{306} &
  306.4 &
  1382.58 &
  \underline{-5.6} &
  0.11 \\
le450\_25b &
450 &
  307 &
  309 &
  309 &
  0 &
  \textbf{307} &
  307 &
  51.43 &
  \textbf{307} &
  307 &
  81.25 &
  \underline{-2} &
  0 \\
miles1500 &
128 &
  797* &
  \textbf{797} &
  797 &
  0 &
  \textbf{797} &
  797 &
  2.04 &
  \textbf{797} &
  797 &
  5.1 &
  0 &
  0 \\
p41 &
116 &
  2688* &
  \textbf{2688} &
  2688 &
  0 &
  \textbf{2688} &
  2688 &
  0.42 &
  \textbf{2688} &
  2688 &
  4.2 &
  0 &
  0 \\ \hline
p42 &
138 &
  2466* &
  \textbf{2466} &
  2466 &
  0.95 &
  \textbf{2466} &
  2466 &
  6.73 &
  \textbf{2466} &
  2466 &
  26.3 &
  0 &
  0 \\
queen10\_10 &
100 &
  162 &
  169 &
  169 &
  7.35 &
  \textbf{162} &
  162 &
  20.91 &
  \textbf{162} &
  162 &
  29.35 &
  \underline{-7} &
  0 \\
queen16\_16 &
256 &
  234 &
  244 &
  244 &
  70.25 &
  240 &
  241.86 &
  873.87 &
  241 &
  241.85 &
  1411 &
  \underline{-2.15} &
  -0.01 \\
r28 &
288 &
  9407* &
  \textbf{9407} &
  9407 &
  0 &
  \textbf{9407} &
  9407 &
  0.31 &
  \textbf{9407} &
  9407 &
  23.2 &
  0 &
  0 \\
r29 &
281 &
  8693* &
  8694 &
  8694 &
  0.2 &
  \textbf{8693} &
  8693 &
  5.51 &
  \textbf{8693} &
  8693 &
  20.85 &
  \underline{-1} &
  0 \\ \hline
r30 &
301 &
  9816* &
  9818 &
  9818 &
  0 &
  \textbf{9816} &
  9816 &
  1.31 &
  \textbf{9816} &
  9816 &
  31 &
  \underline{-2} &
  0 \\
R50\_5gb &
50 &
  135* &
  \textbf{135*} &
  135 &
  384.4 &
  \textbf{135} &
  135 &
  0.21 &
  \textbf{135} &
  135 &
  1.35 &
  0 &
  0 \\
R75\_1gb &
70 &
  70* &
  72 &
  72 &
  1291.1 &
  \textbf{70} &
  70 &
  0.31 &
  \textbf{70} &
  70 &
  1.6 &
  \underline{-2} &
  0 \\
wap01a &
2368 &
  545 &
  603 &
  603 &
  1385.9 &
  552 &
  555.14 &
  2499.01 &
  550 &
  554.7 &
  1512 &
  \underline{-48.3} &
  -0.44 \\
wap02a &
2464 &
  538 &
  591 &
  591 &
  1586.05 &
  544 &
  546.14 &
  1739.2 &
  545 &
  547.25 &
  2867.33 &
  \underline{-43.75} &
  1.11 \\ \hline
  &
 &
   &
  \multicolumn{3}{c|}{6/25} &
  \multicolumn{3}{c|}{14/25} &
  \multicolumn{3}{c|}{14/25} &
   & \\
    \hline
  
\end{tabular}
}
\end{table}

\section{Conclusions and discussion} \label{sec:conclusion}

In this work, we investigated Monte Carlo Tree Search applied to the weighted vertex coloring problem.  We studied different greedy strategies and local searches used for the simulation phase. Our experimental results lead to three conclusions. 

When the instance is large, and when a time limit is imposed, MCTS does not have the time to learn promising areas in the search space and it seems more beneficial to favor more intensification, which can be done in three different ways: (i) by lowering the coefficient which balances the compromise between exploitation and exploration during the selection phase, (ii) by using a dedicated heuristic exploiting the specificity of the problem (grouping in priority the heaviest vertices in the first groups of colors), (iii) and by using a local search procedure to improve the complete solution. For very large instances, we also observed that it may be better to use the local search alone, to avoid too early restarts with the MCTS algorithm, which prevents the local search from improving its score.

Conversely, for small instances, it seems more beneficial to encourage more exploration, to avoid getting stuck in local optima. It can be done, by increasing the coefficient, which balances the compromise between exploitation and exploration, and by using a simulation strategy with more randomness, which favors more exploration of the search tree and also allows a better evaluation of the most promising branches of the MCTS tree. For these small instances, the MCTS algorithm can provide some optimality proofs.

For medium instances, it seems important to find a good compromise between exploration and exploitation. For such instances, coupling the MCTS algorithm with a local search procedure allows finding better solutions, which cannot be reached by the MCTS algorithm or the local search alone.

Other future works could be envisaged. For instance, it could be interesting to use a more adaptive approach to trigger the local search, or to use a machine-learning algorithm to guide the search toward more promising branches of the search tree. 

\section*{Acknowledgements} \label{sec:acknowledgements}

We would like to thank the reviewers for their useful comments. We thank Dr. Wen Sun \cite{Sun_Hao_Lai_Wu_2018}, Dr. Yiyuan Wang, \cite{Wang_Cai_Pan_Li_Yin_2020} and Pr. Bruno Nogueira \cite{Nogueira_Tavares_Maciel_2021} for sharing their codes. This work was granted access to the HPC resources of IDRIS (Grant No. 2020-A0090611887) from GENCI and the Centre Régional de Calcul Intensif des Pays de la Loire.

\bibliographystyle{splncs04}
% \bibliography{biblio}
\bibliography{main.bbl}

\end{document}